\documentclass{article}

    \usepackage[preprint]{tackling_climate_workshop_style}

\usepackage[utf8]{inputenc} 
\usepackage[T1]{fontenc}    
\usepackage{hyperref}       
\usepackage{url}            
\usepackage{booktabs}       
\usepackage{amsfonts}       
\usepackage{nicefrac}       
\usepackage{microtype}      
\usepackage{graphicx}       
\usepackage{float}          
\usepackage{subcaption}     
\usepackage{svg}            
\usepackage{booktabs}       
\usepackage{multirow}

\title{The Garbage Dataset (GD): A Multi-Class Image Benchmark for Automated Waste Segregation}

\author{%
  Suman Kunwar \\
  DWaste, USA\\
  \texttt{sumn2u@gmail.com} \\
}

\begin{document}

\maketitle

\begin{abstract}
This study introduces the Garbage Dataset (GD), a publicly available image dataset designed to advance automated waste segregation through machine learning and computer vision. It is a diverse dataset that covers 10 categories of common household waste: metal, glass, biological, paper, battery, trash, cardboard, shoes, clothes, and plastic. The dataset comprises 12,259 labeled images collected through multiple methods, including  the DWaste mobile app and curated web sources. The methods included rigorous validation through checksums and outlier detection, analysis of class imbalance and visual separability through PCA/t-SNE, and assessment of background complexity using entropy and saliency measures. The dataset was benchmarked using state-of-the-art deep learning models (EfficientNetV2M, EfficientNetV2S, MobileNet, ResNet50, ResNet101) evaluated on performance metrics and operational carbon emissions. The results of the experiment indicate that EfficientNetV2S achieved the highest performance with a accuracy of 95.13\% and an F1-score of 0.95 with moderate carbon cost. Analysis revealed inherent dataset characteristics including class imbalance, a skew toward high-outlier classes (plastic, cardboard, paper), and brightness variations that require consideration. The main conclusion is that GD provides a valuable real-world benchmark for waste classification research while highlighting important challenges such as class imbalance, background complexity, and environmental trade-offs in model selection that must be addressed for practical deployment. The dataset is publicly released to support further research in environmental sustainability applications.
\end{abstract}


\section{Introduction}

Effective waste segregation is a critical bottleneck in global recycling systems. With solid waste generation projected to increase by 73\% to 3.88 billion tons annually by 2050 \citep{kaza_more_2021}, and the US dumping over two-thirds of its waste despite high per-capita generation \citep{us_epa_national_2017}, automated sorting technologies are urgently needed. Computer vision offers a promising solution, but its development is constrained by the availability of robust, large-scale, and well-characterized image datasets.
A review of the relevant literature reveals several existing waste image datasets, each with specific limitations. TrashNet \citep{thung_garythung/trashnet_2026} is widely used but lacks class diversity. TACO \citep{proenca_taco:_2020} focuses on waste in natural environments, while UAVVaste \citep{kraft_autonomous_2021} provides an aerial perspective. SpotGarbage-GINI \citep{mittal_spotgarbage:_2016}  and Trashbox \citep{kumsetty_trashbox:_2022} are sourced from refined web searches, and there are specialized datasets for items such as cigarette butts \citep{noauthor_cigarette_2018}, plastics \citep{choras_pet_2018} or marine debris \citep{hong_trashcan:_2020, fulton_robotic_2018}. A significant gap remains for a large-scale, multi-class dataset focused on common household recyclables and trash, curated with detailed characterization to inform model development and highlight inherent data challenges.

Various deep learning approaches have been used for waste classification. MobileNet and ResNet50 are common benchmarks \citep{poudel_classification_2022}, and ResNet variants are frequently used \citep{al-mashhadani_waste_2023}. To provide a comprehensive evaluation, we benchmark our dataset with multiple model families selected for their architectural innovations and practical relevance. We include ResNet50 and ResNet101, which utilize skip connections to address vanishing gradients in deeper networks, enhancing optimization without a proportional computational increase \citep{he_deep_2016}. MobileNet is incorporated for its efficiency on mobile and edge devices through depthwise separable convolutions \citep{howard_mobilenets:_2017}. Finally, we evaluated the EfficientNetV2 models (specifically the S and M variants) for their superior training speed and parameter efficiency compared to earlier architectures \citep{tan_efficientnetv2:_2021}.

This work is based on the idea that a comprehensively curated and analyzed dataset will not only provide a superior benchmark for waste classification models, but will also illuminate critical data-centric factors, such as class imbalance, background complexity, and visual separability that fundamentally impact real-world performance. To test this, we present the GD and adopt a multi-faceted approach: (1) multi-source data collection and rigorous validation, (2) in-depth statistical and visual analysis to quantify dataset properties, and (3) extensive benchmarking using the aforementioned deep learning architectures using the transfer learning approach. The models are evaluated in both original images and standardized versions (resized to 256×256 and 384×384 pixels) using performance metrics (accuracy, recall, F1-score), training time, and operational carbon emissions.

This approach is justified by the need to move beyond simple model accuracy comparisons and toward an understanding of how dataset attributes influence practical outcomes, including environmental impact. The principal results confirm that GD is a challenging  real-world benchmark where the choice of model architecture significantly outweighs the benefit of simple image resizing, and where the highest accuracy (95.13\% with EfficientNetV2S) comes with a measurable carbon cost. The main conclusions are that GD fills an important resource gap and that successful waste classification models must be co-designed with consideration of the data inherent biases and the environmental footprint of the training process.

\section{The Garbage Dataset (GD)}
\subsection{Data Collection and Curation}

\begin{figure}[H]
  \centering
  \includegraphics[trim={10 0 10 10}, clip, width=\textwidth]{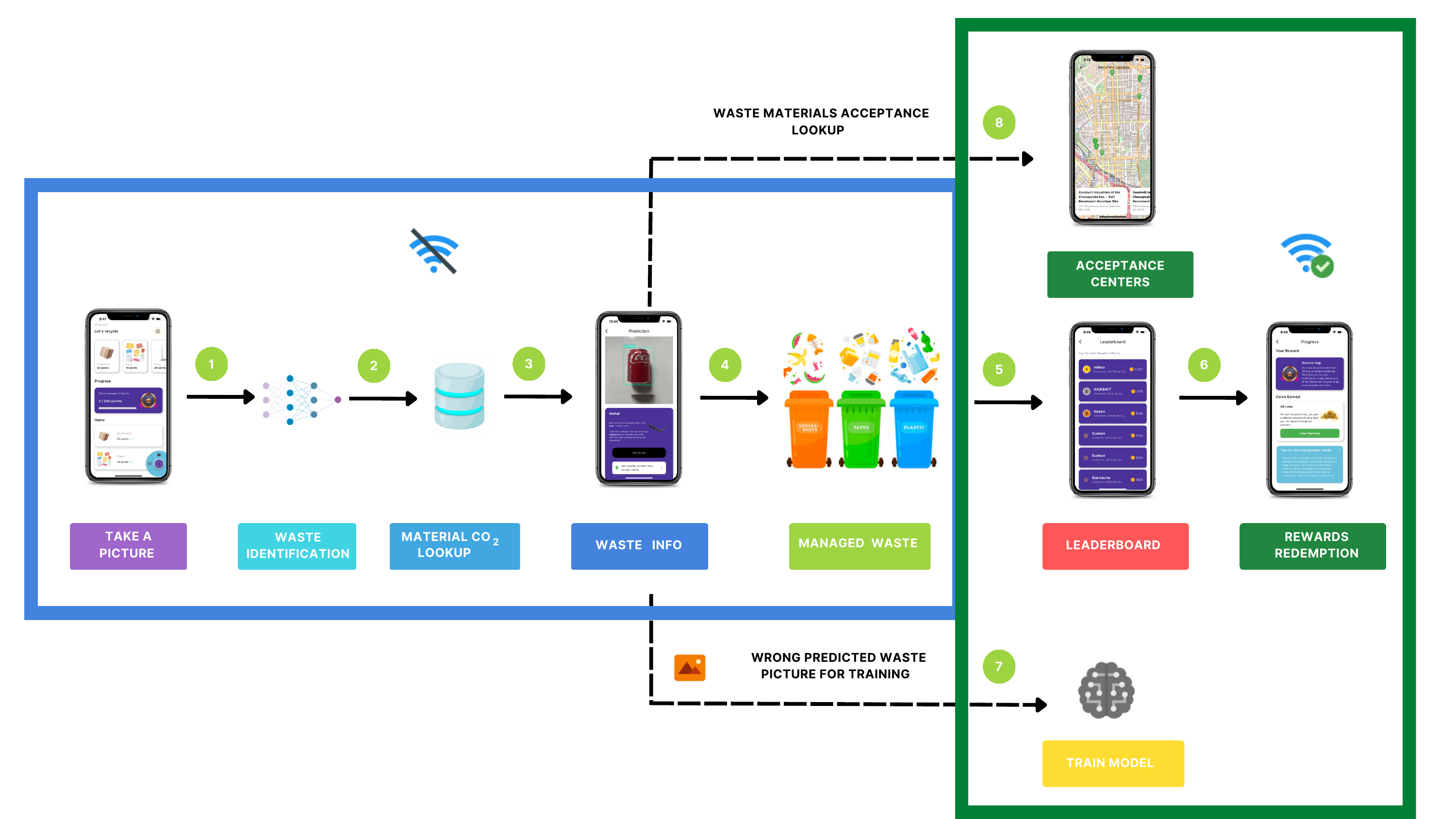}
  \caption{Interface and workflow of the DWaste mobile application for field data collection}
  \label{fig:dwaste_workflow}
\end{figure}

The GD was compiled using a multi-method approach to ensure diversity and real-world applicability. The collection involved: (i) capture images of waste items against various backgrounds using mobile cameras and the dedicated DWaste mobile application; (ii) obtain and curate images from publicly available repositories and web scraping; and (iii) accept community submissions. This methodology ensures that the dataset reflects the heterogeneity of real-world waste scenarios. The initial dataset contains a total of 20,212 images, spanning several classes shown in Figure \ref{fig:dwaste_workflow}. These images passed through several cleaning phases, where images are checked for duplication, integrity, and copyright checks. The collected images went through two image hashing approaches: MD5 hashing and Perceptual hashing. 

\begin{figure}[H]
  \centering
  \includegraphics[width=0.96\textwidth]{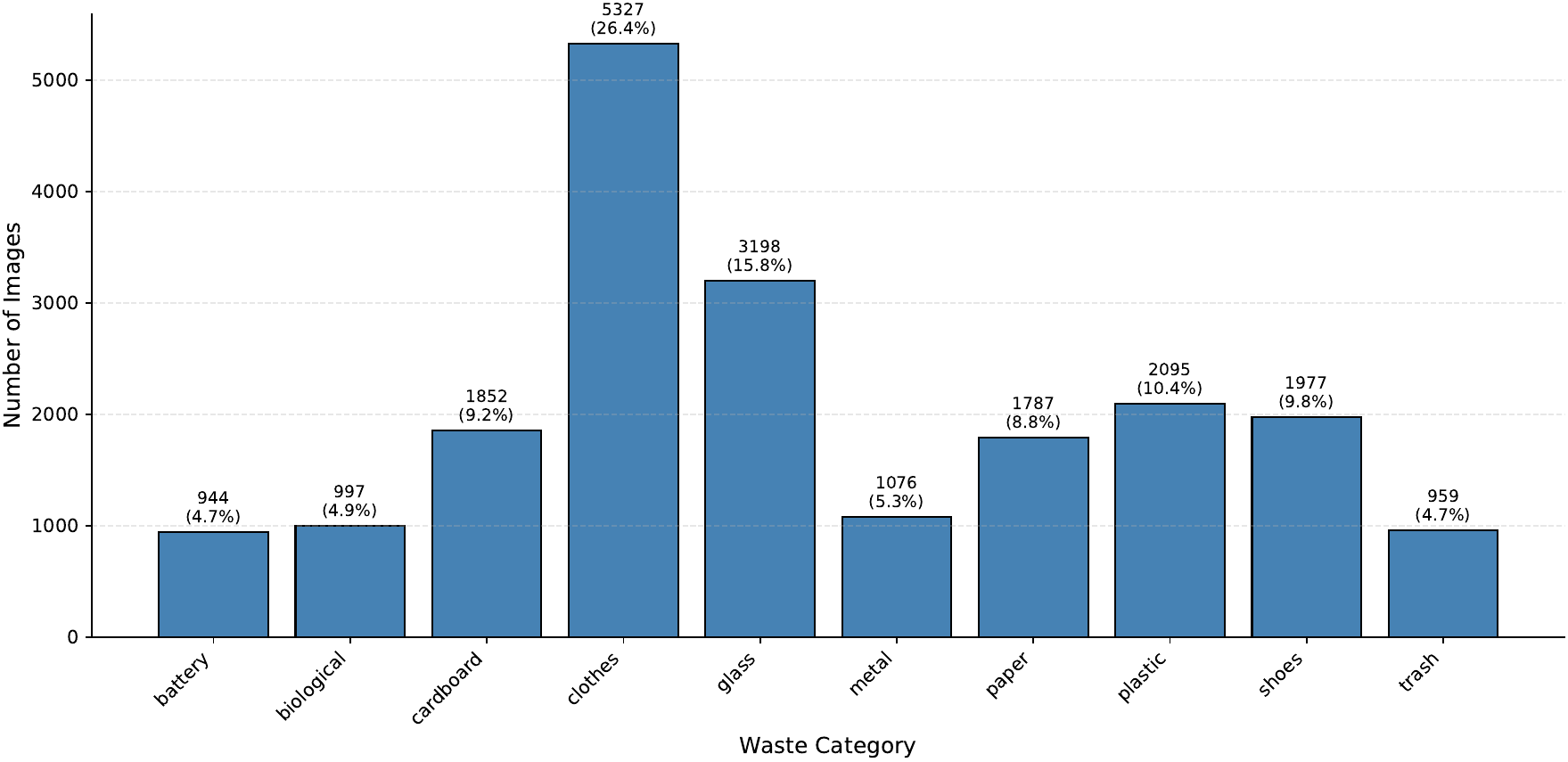}
  \caption{Class distribution of the collected images}
  \label{fig:class_distribution}
\end{figure}
 
MD5 hashing was used to verify the  integrity  of the file \citep{shaikh_image_2024} along with the exact duplication, while perceptual hashing was used for images close to duplicate \citep{zhou_survey_2025}, as shown in Algorithm \ref{alg:duplicate_detection_short}.

\begin{algorithm}[H]
\caption{Duplicate and Near-Duplicate Detection using Hashing}
\label{alg:duplicate_detection_short}
\begin{algorithmic}[1]
\REQUIRE Set of image paths $\mathcal{I}$, Hamming threshold $\tau = 5$
\ENSURE Lists $\mathcal{D}_{exact}$, $\mathcal{D}_{near}$
\STATE Initialize dictionaries $\mathcal{H}_{MD5}$, $\mathcal{H}_{pHash}$ and lists $\mathcal{D}_{exact}$, $\mathcal{D}_{near}$
\FOR{each $I_i \in \mathcal{I}$}
    \STATE $h_{md5} \gets \text{MD5}(I_i)$
    \IF{$h_{md5} \in \mathcal{H}_{MD5}$}
        \STATE Append $(I_i, \mathcal{H}_{MD5}[h_{md5}])$ to $\mathcal{D}_{exact}$
    \ELSE
        \STATE $\mathcal{H}_{MD5}[h_{md5}] \gets I_i$
        \STATE Resize $I_i$ to $32\times32$; $h_p \gets \text{pHash}(I_i)$
        \FOR{each $(h_p^j, I_j) \in \mathcal{H}_{pHash}$}
            \IF{Hamming$(h_p, h_p^j) < \tau$}
                \STATE Append $(I_i, I_j)$ to $\mathcal{D}_{near}$
                \STATE \textbf{break}
            \ENDIF
        \ENDFOR
        \IF{no match found} \STATE $\mathcal{H}_{pHash}[h_p] \gets I_i$ \ENDIF
    \ENDIF
\ENDFOR
\RETURN $\mathcal{D}_{exact}, \mathcal{D}_{near}$
\end{algorithmic}
\end{algorithm}

Here, an MD5 hash is computed for every image to identify exact duplicates; if a  hash matches a previously seen image, the pair is recorded. If it is unique, it then computes a perceptual hash (pHash), resizing the image to 32×32 beforehand. It compares this pHash with all previously stored pHashes using the Hamming distance; if the distance is below a threshold (5), the images are considered near‑duplicates and the pair is recorded. If no near-‑duplicate is found, the pHash is stored for future comparisons. The algorithm outputs exact duplicate pairs and close to duplicate pairs. From image hashing, an exact duplicate and 1,360 near duplicate images were found which were removed. 

A total of 720 transparent images were removed from the dataset during preprocessing, as recent studies show that models struggle with transparent objects \citep{wen_seeing_2025}. In addition, 20 non-RGB images (with color modes P, CMYK, and L) were removed to ensure a uniform input representation and more stable results. Different color spaces provide different numerical representations, which can significantly affect model performance \citep{jawahar_colornet:_2019}.

Some images contained watermarks, were copyrighted, or had text on the edges. These watermark images are likely to reduce model performance \citep{yu_robust_2025}.  TrustMark was used to detect watermarks \citep{bui_trustmark:_2023}, identifying 493 watermark images. In addition, manual verification was performed to remove copyrighted images, those with text on the edges, and watermarked images. This process reduced the dataset from an initial 18,114 images to 12,259. To achieve a more balanced class distribution, classes such as glass and clothes were downsampled.

\begin{figure}[H]
  \centering
  \includegraphics[width=0.95\textwidth]{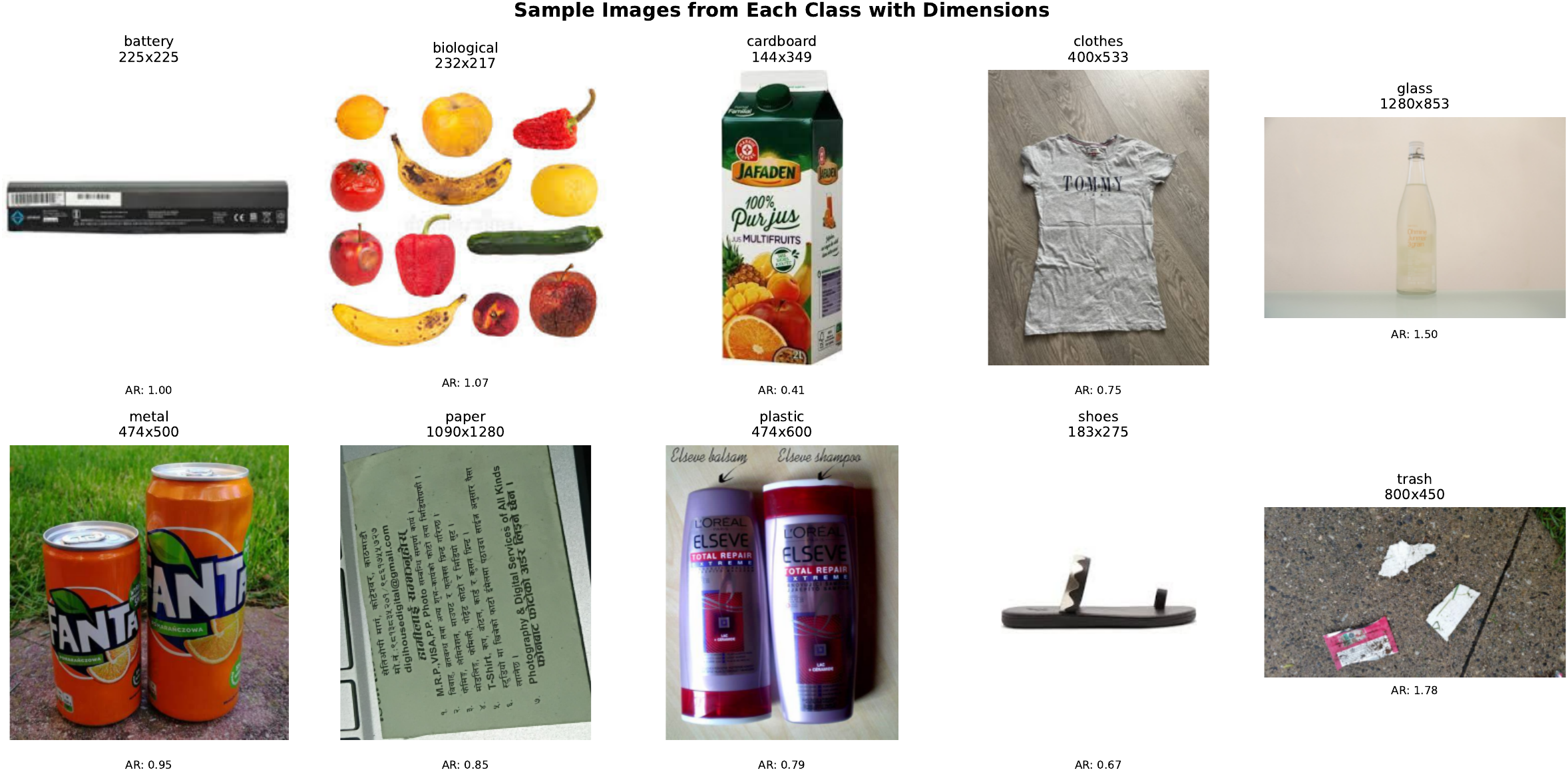}
  \caption{Example images from the dataset, illustrating variation in object type, scene context, and image quality}
  \label{fig:sample_images}
\end{figure}

Within a given image, multiple items may appear, but all such items belong to the same annotated class. The dataset captures a wide range of real-world variability, including indoor and outdoor environments, diverse lighting conditions, various background contexts, and physical deformations such as crushed, crumpled, or bent items. This ensures that the images reflect the heterogeneity and challenging conditions encountered in practical waste classification scenarios. The dataset is organized in a flat directory structure, with a main folder containing subdirectories for each class, each housing the respective JPEG files. This structure facilitates straightforward integration with standard machine learning pipelines. Figure \ref{fig:sample_images} illustrates sample images, and Figure \ref{fig:dwaste_workflow} illustrates the image collection from the DWaste mobile app.

\newpage
\subsection{Dataset Statistics and Structure}
\begin{figure}[H]
  \centering
  \includegraphics[width=\textwidth]{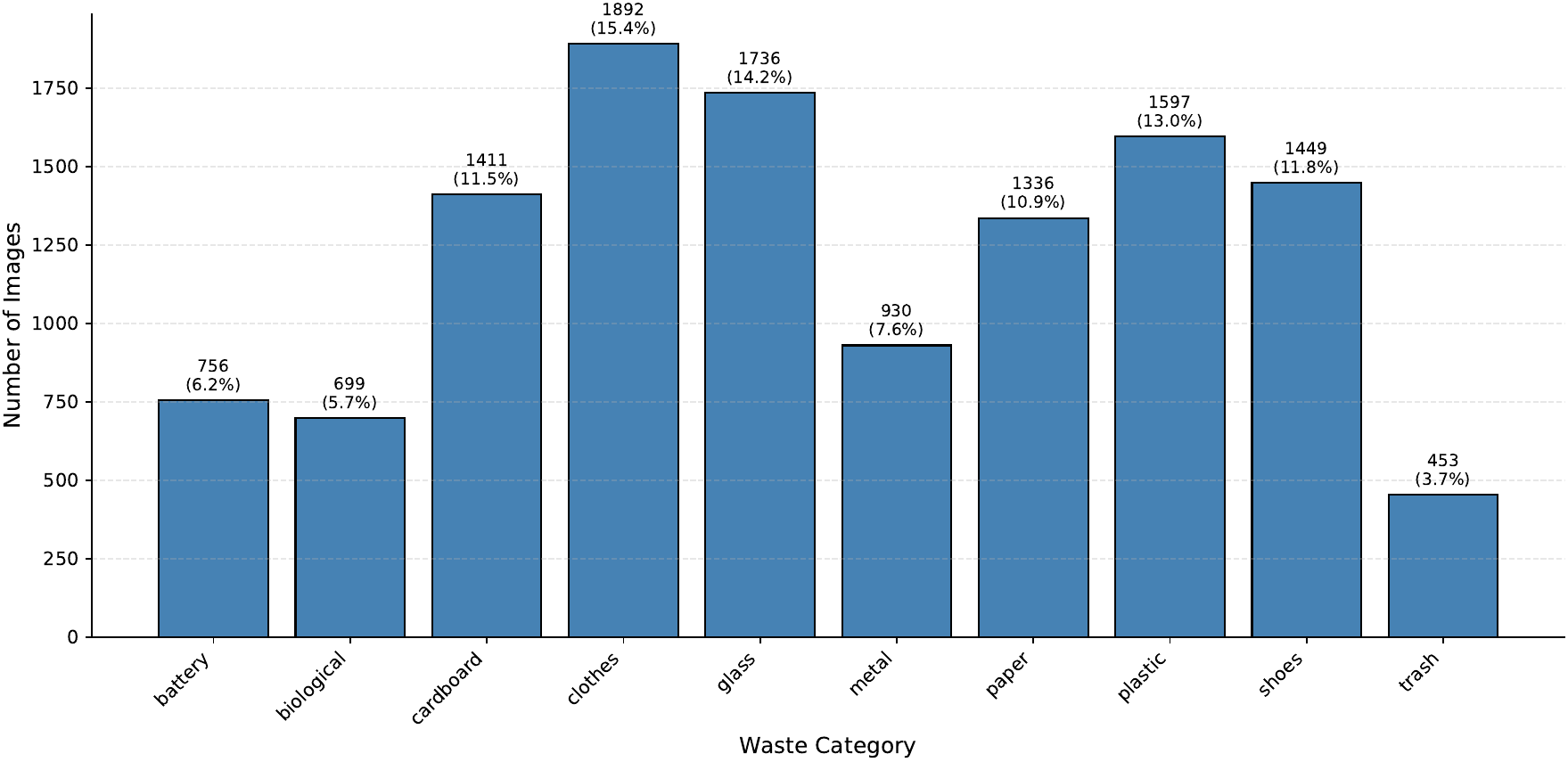}
  \caption{Class distribution of the cleaned dataset}
  \label{fig:class_distribution_cleaned}
\end{figure}

The dataset comprises 12,259 JPEG images totaling approximately 1.12 GB. A significant class imbalance exists: for instance, the "Glass" category contains 1,736 images, which is nearly four times the 453 images in the "Trash" category which can be seen in Figure \ref{fig:class_distribution_cleaned}. This imbalance may bias models toward majority classes without corrective strategies such as oversampling, undersampling, or class-weighted loss functions.

As images are collected from multiple sources, there is notable cross-source variation.  Images captured via the DWaste mobile application tend to exhibit consistent resolution, as users are guided to photograph items against neutral backgrounds and are saved with fixed resolution size. In contrast, web-scraped images vary widely in resolution, compression artifacts, and background complexity, ranging from clean studio setups to cluttered real-world scenes. 

\begin{figure}[H]
  \centering
  \includegraphics[width=0.95\textwidth]{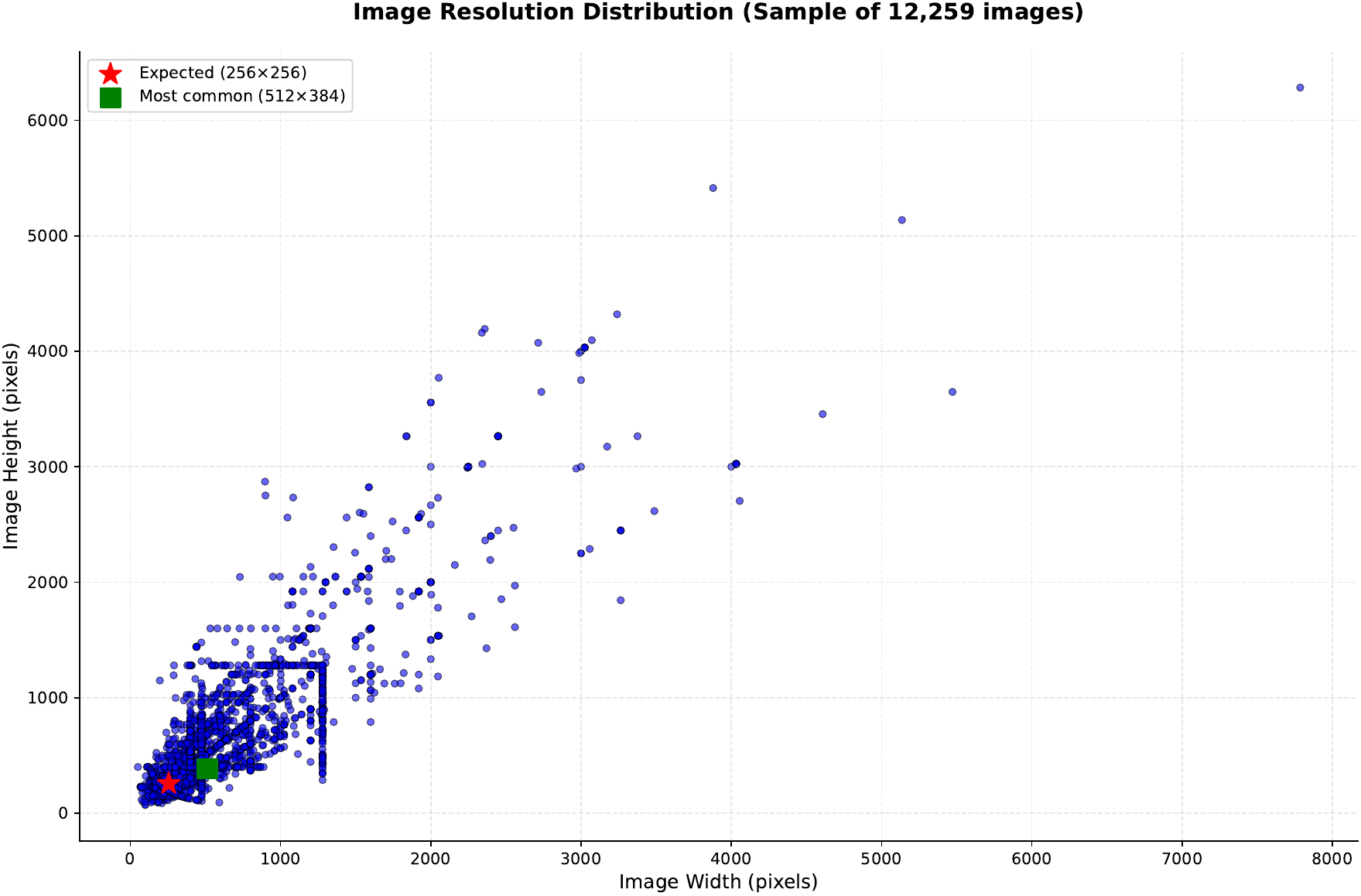}
  \caption{Distribution of image resolutions}
  \label{fig:image_resolutions}
\end{figure}

Community submissions fall between these extremes, often featuring diverse indoor and outdoor environments with uncontrolled lighting and occlusion, as images are collected from recycling centers, parks, and bins. The differences in image size can be seen in Figure \ref{fig:image_resolutions}. These source differences create potential domain shifts in  brightness, color, sharpness, and background context. Although this heterogeneity poses challenges for models trained on homogeneous data \citep{guo_survey_2023}, it is precisely this variability that makes the dataset a robust benchmark for real-world generalization \citep{xiao_generalization_2024}. Researchers can leverage these cross-source distinctions to study the domain adaptation and robustness of the model, a key to deploying waste classification systems in uncontrolled environments. They are advised to employ standard preprocessing and augmentation, as research confirms that image size \citep{hieu_impact_2024} and resolution \citep{du_impact_2025} significantly affect model performance.
\newpage

\begin{figure}[H]
  \centering
  \includegraphics[trim={10 10 10 10}, clip, width=\textwidth]{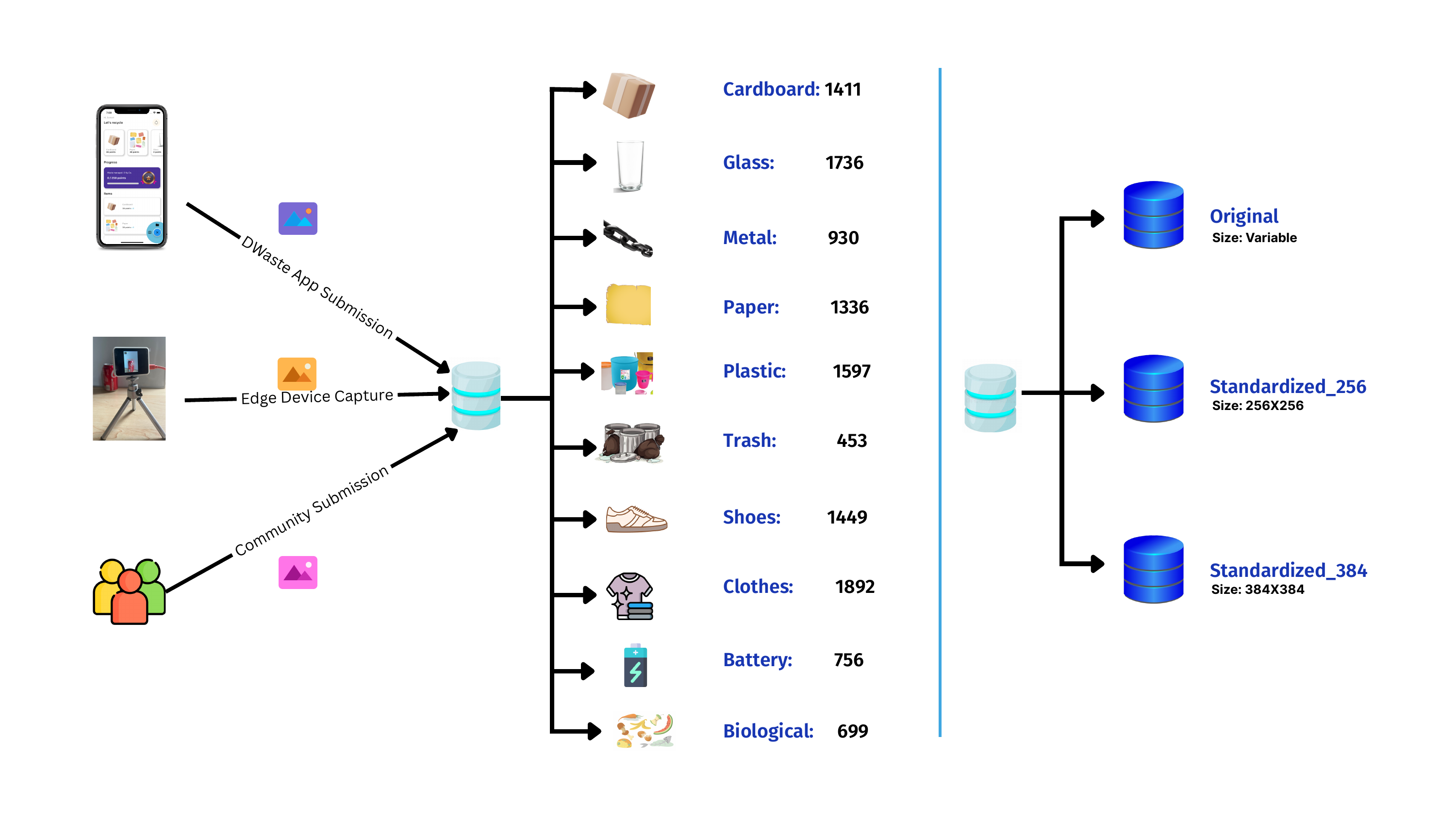}
  \caption{Dataset summary}
  \label{fig:dataset_summary}
\end{figure}

To accommodate the input dimensions of the standard model, we provide three versions of the dataset: Original, Standardized\_256 \citep{ghosh_reshaping_2019}, and Standardized\_384. Normalized versions are created by resizing the images to 256×256 and 384×384 pixels, respectively, using padding to preserve the aspect ratio. Figure \ref{fig:dataset_summary} shows the data sources, the image counts for each class, and its standardization. The dataset is publicly available on Kaggle\footnote{\url{https://www.kaggle.com/datasets/sumn2u/garbage-classification-v2}}.

\subsection{Dataset Quality and Validation}
\begin{figure}[H]
  \centering
  \includegraphics[width=0.95\textwidth]{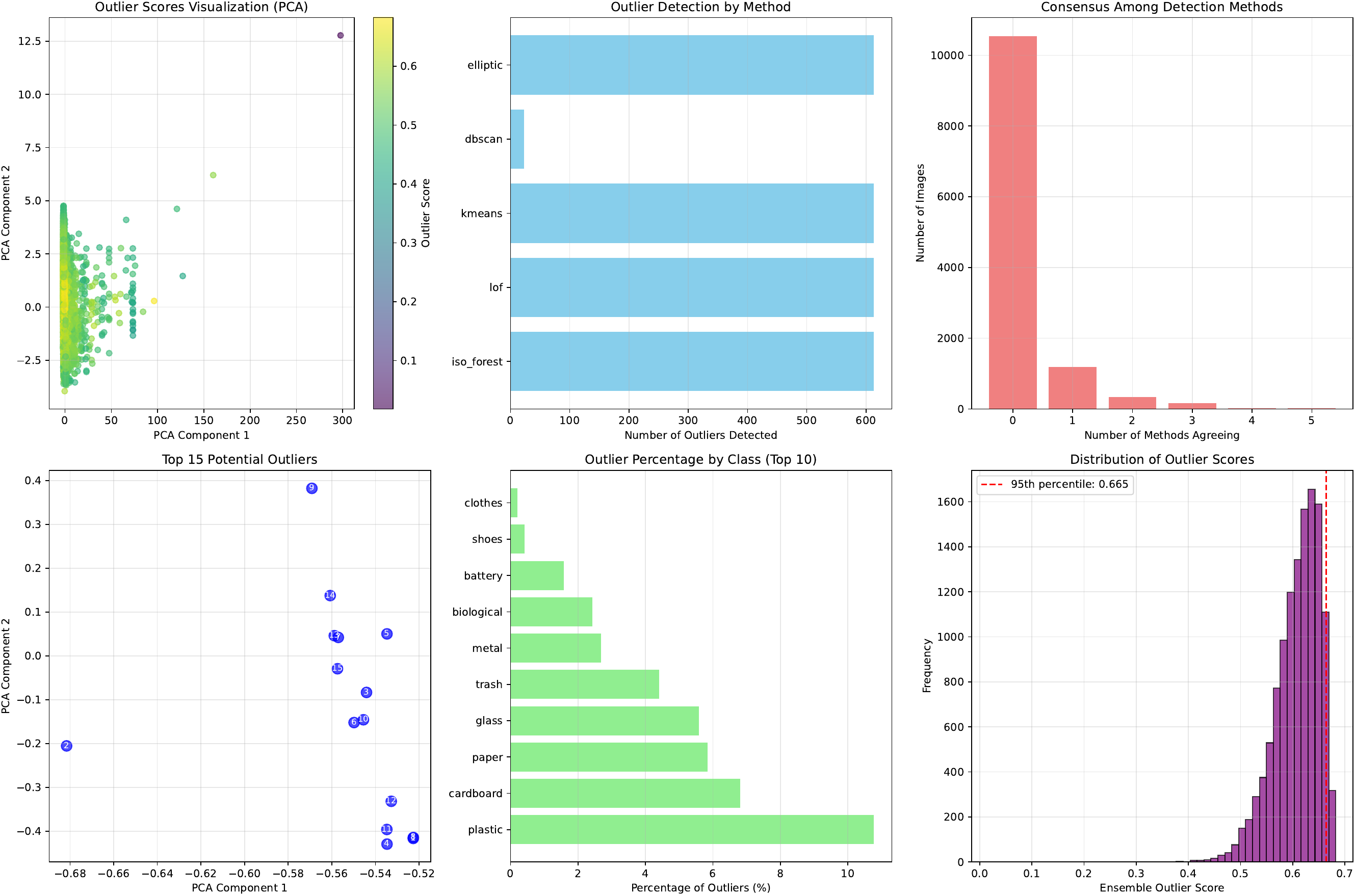}
  \caption{Outliers exploration}
  \label{fig:outliners_exploration}
\end{figure}

Data integrity was verified using checksums and Cyclic Redundancy Check (CRC) \citep{liu_checksum_2009}, confirming that there were no transmission or storage errors. Each image was manually annotated into one of ten predefined categories and verified by at least three volunteers. We ensured that all labels are accurate and that no personally identifiable or copyrighted images are included. The images collected through the community and the app were obtained with user permission, and a mechanism is provided for users to request the removal of their images if desired. Outlier detection detected 4.3\% (527 images) as anomalies. The distribution was very uneven: plastic (10.8\%), cardboard (6.8\%) and paper (5.8\%) had the highest rates, while clothes (0.2\%) and shoes (0.4\%) showed exceptional consistency. This indicates that while the dataset is mostly clean, a targeted review of high-outlier classes could further enhance the quality. The visual clue of outliers can be seen in Figure \ref{fig:outliners_exploration}.

\section{Dataset Analysis}

\subsection{Background and Foreground Analysis}
\begin{figure}[H]
  \centering
  \includegraphics[width=0.95\textwidth]{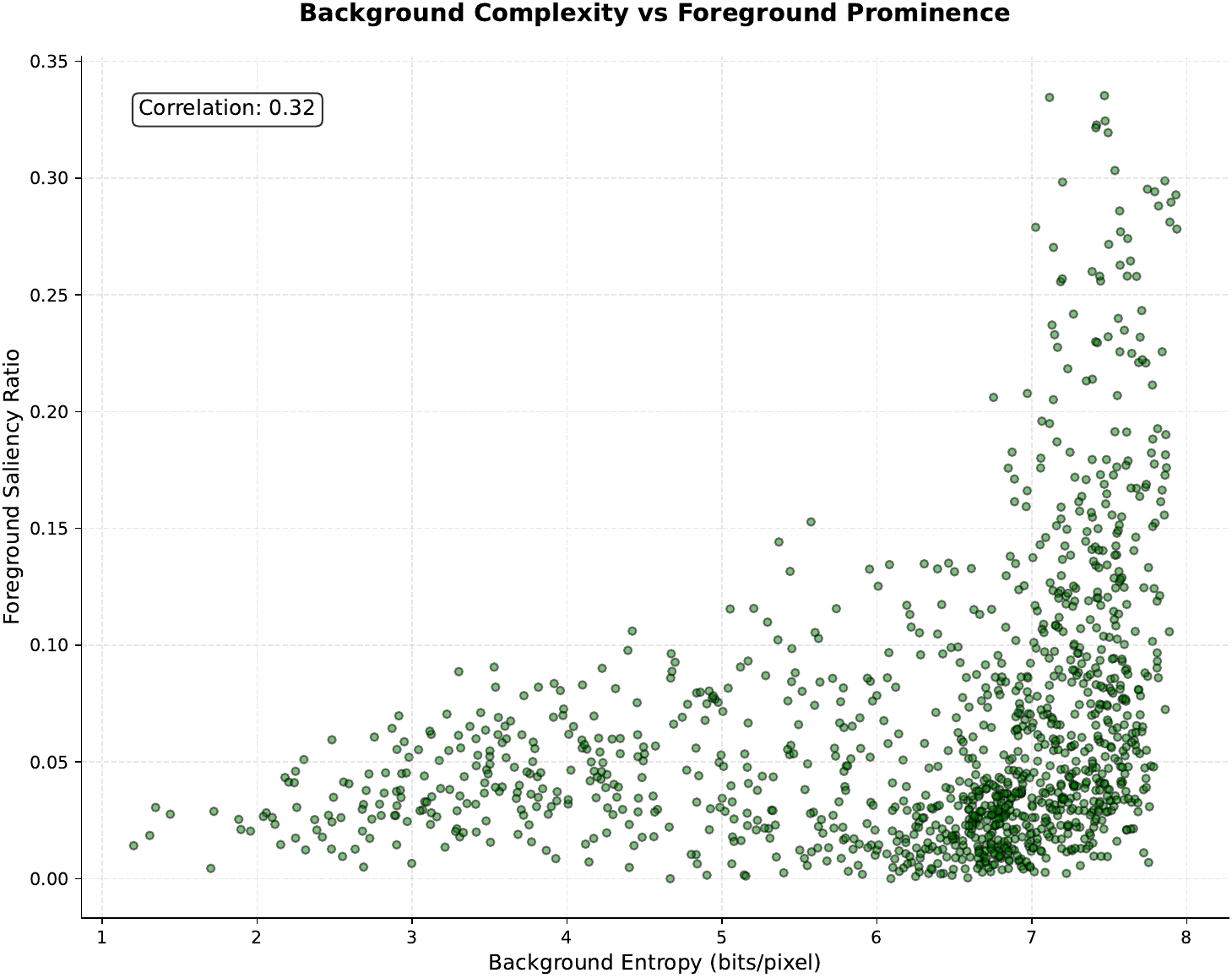}
  \caption{Background entropy vs foreground saliency ratio of images}
  \label{fig:saliency}
\end{figure}

The complexity of the background significantly impacts the accuracy of the classification \citep{sielemann_measuring_2025}. We analyze this by mapping the foreground saliency ratio to the Background Shannon Entropy, where a higher entropy indicates greater visual complexity \citep{ma_combined_2025}. The resulting distribution for the dataset is provided in Figure \ref{fig:saliency}.
\newpage

Analysis of all 12,259 images revealed high background complexity, with mean Shannon entropy of 6.3 ± 1.5 bits/pixel. This is compounded by a low mean Foreground Saliency Ratio of 0.06 ± 0.06, indicating visually dominant backgrounds. Manual categorization of a 500-image sample found that the  backgrounds consisted primarily  of indoor floors (35\%), outdoor ground (28\%), and tables/surfaces (22\%).

Saliency detection can improve computational efficiency by focusing on informative regions, reducing search space, and false positives in complex scenes \citep{qiu_research_2024}. Furthermore, lighting conditions pose challenges; our analysis indicates a brightness factor skewed toward over-exposure (mean: 1.35), which may impact model performance \citep{rodriguez-rodriguez_impact_2024, wang_contrast_2023}.

\subsection{Class Imbalance and Visual Separability}
\begin{figure}[H]
  \centering
  \includegraphics[width=0.95\textwidth]{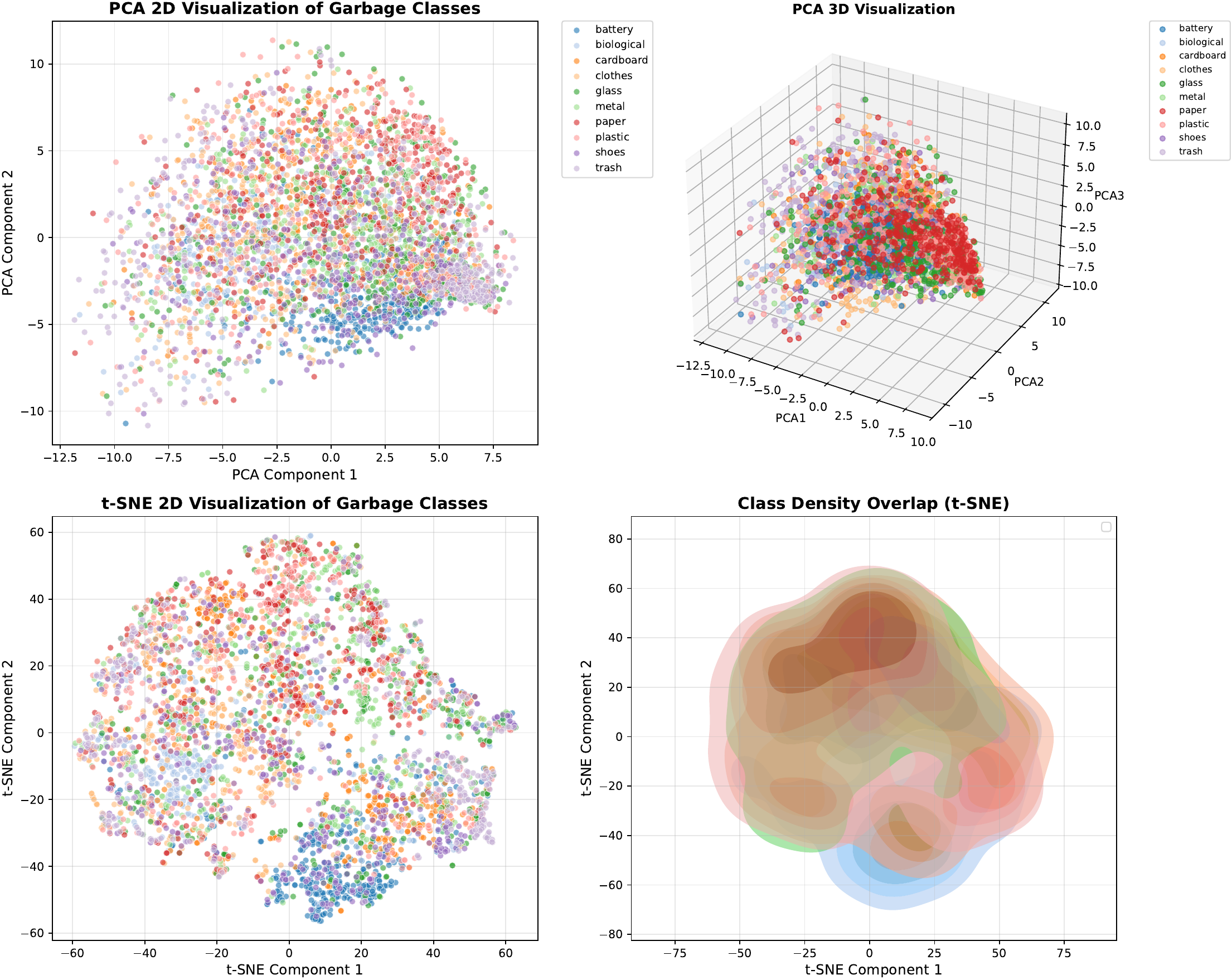}
  \caption{Visualization of PCA/t-SNE of the images}
  \label{fig:visual_separability}
\end{figure}

We assessed visual separability using Principal Component Analysis (PCA) and t-Distributed Stochastic Neighbor Embedding (t-SNE) \citep{arora_analysis_2018} in a balanced subset (400 images per class). PCA revealed poor linear separability, capturing only 31.56\% of variance. In contrast, t-SNE showed moderate clustering with a separability ratio of 1.24 (avg distance to the same class: 1.0922, avg distance to a different class: 1.3498) as shown in Figure \ref{fig:visual_separability}. \newpage

The most challenging classes, indicated by high inter-class overlap, were shoes (0.788), glass (0.777), metal (0.759), and plastic (0.695). Paper and plastic emerged as the most frequently confused pair, with a recorded centroid distance of 4.12. Contrary to initial assumptions, biological (0.567) and trash (0.555) were among the more distinct categories. This complexity confirms the suitability of the dataset not only for image classification but also for research in object detection, data augmentation, transfer learning, and few-shot learning.

\section{Benchmark Experiments}
\subsection{Experimental Setup}

The dataset was divided into training sets (80\%), validation sets (10\%), and test sets (10\%) for experimental purposes. To mitigate class imbalance, random undersampling was applied to the majority of classes during training. All benchmarked models (EfficientNetV2M, EfficientNetV2S, MobileNet, ResNet50, ResNet101) utilized transfer learning from ImageNet weights. Each model was evaluated in three versions of the dataset (Original, Standardized\_256, Standardized\_384) using a NVIDIA Tesla T4x2 GPU in Kaggle. Performance was measured over 20 epochs using accuracy, recall, and F1-score. Training time and operational carbon emissions were tracked using Code Carbon \citep{benoit_courty_2024_11171501}.

\subsection{Results and Discussion}

The benchmark results are shown in Table \ref{tab:models_benchmark} reveals EfficientNetV2S as the best performing model, achieving 95.13\% accuracy and a 0.95 F1 score with a training time of 6,058 seconds. In contrast, the fastest model, MobileNet\_256, trains in 2,480 seconds but sacrifices 28 percentage points in accuracy, scoring only 67.88\%. Increasing the input resolution to 384x384 generally provided minimal accuracy gains (under 1\%) while significantly increasing the computational cost. Notably, EfficientNetV2 architectures consistently outperformed ResNet models, with EfficientNetV2S exceeding ResNet101's accuracy by 2.36 percentage points despite similar training times.

Further analysis reveals that the minority `trash' class (3.7\% of the data) consistently yielded the lowest F1-scores across all models. Performance dropped to 0.40 on MobileNetV2 and peaked at only 0.90 on the best-performing model. These results confirm that without targeted interventions, such as class weighting, models remain biased against under-represented yet critically important categories. Visual overlap between the `paper and `plastic' classes remained a challenge, with ResNet101-TL-v2-384 producing identical F1-scores (0.88) for each. Similarly, recall for the `metal' category fluctuated wildly (0.49–0.96) depending on the model used. These results indicate that scaling model capacity is not enough; addressing visual ambiguity requires data-centric solutions like specialized augmentation or tailored architectures.

\begin{table}[htbp]
\caption{Comparative performance of waste classification models.}
\label{tab:models_benchmark}
\small
\begin{tabular*}{\textwidth}{@{\extracolsep{\fill}}lccccccc@{}}
\hline
Model & Time & Accuracy & Recall & F1 Score &
\multicolumn{3}{c}{CO$_2$ Emissions} \\
\cline{6-8}
 &  &  &  &  & Prepare & Develop & Deploy \\
\hline
EN-V2M       & 7137.81 & 94.15 & 0.93 & 0.93 & 0.0011 & 0.1416 & 0.1444 \\
EN-V2M-256   & 6169.28 & 93.61 & 0.93 & 0.93 & 0.0008 & 0.0765 & 0.0784 \\
EN-V2M-384   & 8084.41 & 94.58 & 0.94 & 0.94 & 0.0011 & 0.1583 & 0.1614 \\
EN-V2S       & 6057.87 & 95.13 & 0.95 & 0.95 & 0.0011 & 0.1180 & 0.1204 \\
EN-V2S-256   & 5908.89 & 93.77 & 0.94 & 0.93 & 0.0015 & 0.1147 & 0.1175 \\
EN-V2S-384   & 4044.54 & 94.50 & 0.94 & 0.94 & 0.0010 & 0.0785 & 0.0807 \\
MN           & 2420.15 & 66.94 & 0.65 & 0.65 & 0.0010 & 0.0475 & 0.0493 \\
MN-256       & 2480.46 & 67.88 & 0.66 & 0.66 & 0.0002 & 0.0149 & 0.0153 \\
MN-384       & 2536.73 & 70.39 & 0.68 & 0.68 & 0.0002 & 0.0152 & 0.0156 \\
RN50         & 5228.93 & 92.61 & 0.92 & 0.92 & 0.0015 & 0.1019 & 0.1045 \\
RN50-256     & 5830.07 & 91.91 & 0.91 & 0.91 & 0.0003 & 0.0349 & 0.0356 \\
RN50-384     & 5643.32 & 92.39 & 0.92 & 0.92 & 0.0012 & 0.1115 & 0.1140 \\
RN101        & 7331.92 & 92.77 & 0.93 & 0.93 & 0.0007 & 0.0906 & 0.0924 \\
RN101-256    & 5147.28 & 92.31 & 0.92 & 0.92 & 0.0017 & 0.1008 & 0.1040 \\
RN101-384    & 7445.92 & 92.64 & 0.93 & 0.92 & 0.0003 & 0.0455 & 0.0463 \\
\hline
\end{tabular*}

\vspace{2mm}
\footnotesize
\textbf{Abbreviations:}
EN-V2M = EfficientNetV2M,
EN-V2S = EfficientNetV2S,
MN = MobileNet,
RN50 = ResNet50,
RN101 = ResNet101.
\end{table}

For rapid iteration, EfficientNetV2S\_384 offers a compelling alternative, offering 94.50\% accuracy with relatively low carbon emissions and a fast training time of 4,045 seconds. The results underscore that model architecture selection has a much greater impact on performance than simply increasing input size. The experimental code for reproducing all analyzes and benchmarks is available on GitHub:\footnote{\url{https://github.com/sumn2u/garbage-dataset-experiments}}.

\begin{figure}[H]
  \centering
  \includegraphics[width=0.95\textwidth]{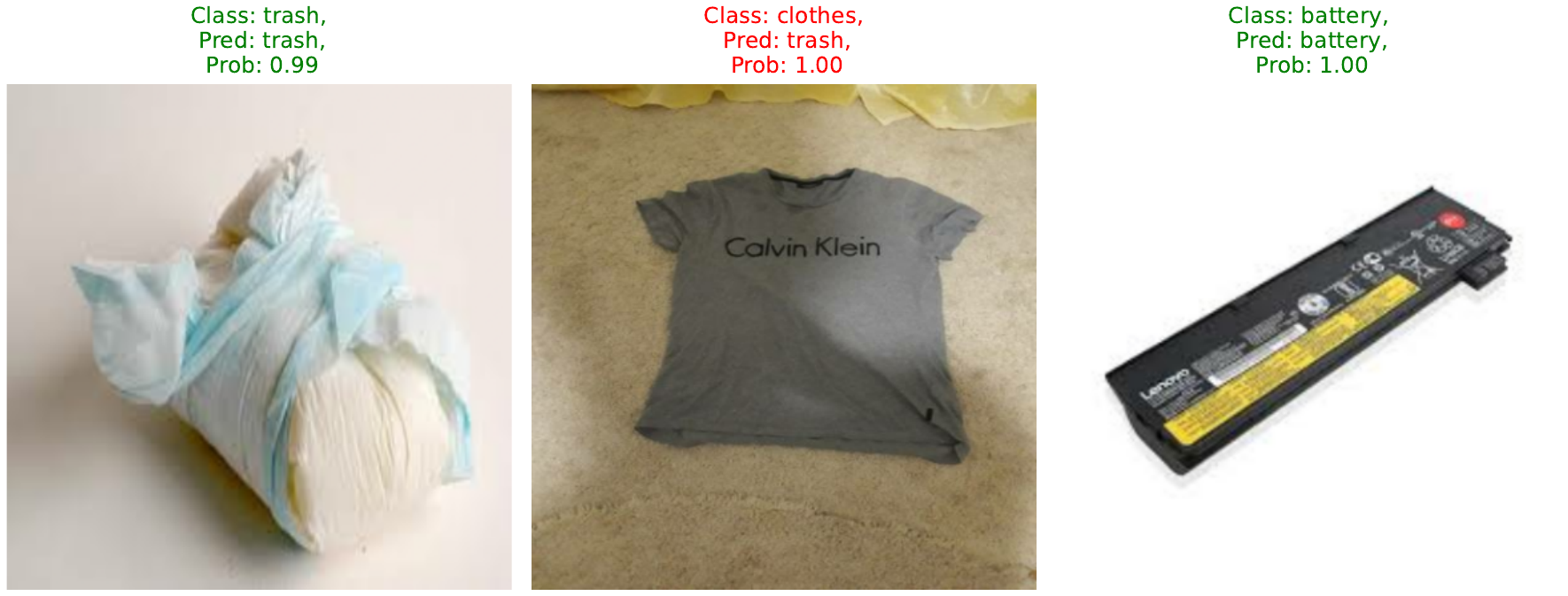}
  \caption{Predictions on sample image with actual size using EfficientNetV2M model}
  \label{fig:inference_example}
\end{figure}

Model selection is also an environmental decision. The least accurate models (MobileNet variants at ~66\%) are the greenest, while the most accurate model (EfficientNetV2S) has a moderate carbon cost. For balanced projects, EfficientNetV2S\_384 offers a near-optimal green alternative, cutting carbon emissions dramatically with a minimal accuracy penalty.  Figure \ref{fig:inference_example} demonstrates EfficientNetV2M predictions where it correctly predicted 2 out of 3 images. By establishing a baseline using undersampling without augmentation, this study highlights the inherent challenges of the raw dataset.

\section{Conclusion}
This study presented GD, a comprehensive multi-source image collection designed to advance automated waste segregation. Our main findings demonstrate that GD provides a challenging real-world benchmark characterized by: (1) significant class imbalance, (2) high visual complexity driven by background and lighting variations, and (3) distinct yet overlapping clusters as revealed by t-SNE analysis. These conclusions directly address the problem outlined in the introduction, the lack of a large-scale, well-characterized dataset for household waste classification. The evidence from our systematic analysis and benchmark experiments confirms that while state-of-the-art models like EfficientNetV2S can achieve high accuracy (95.13\%), performance is fundamentally constrained by the dataset's inherent properties, particularly class imbalance and background noise.

The implications of this work are both practical and methodological. For practical applications, GD serves as a vital resource for the development of robust waste-sorting systems in recycling facilities, public spaces, and educational tools. Methodologically, our work underscores that achieving real-world robustness requires moving beyond optimizing model architecture alone; it necessitates explicit handling of data-centric challenges such as imbalance, background complexity, and environmental cost. The inclusion of carbon emission metrics provides a crucial dimension for sustainable AI development in environmental applications.

Future work should focus on addressing the identified limitations through advanced augmentation, imbalance correction techniques, and the development of models that are accurate and computationally efficient. The release of GD aims to catalyze such research, contributing to scalable and sustainable solutions to global waste management challenges.

\begin{ack}
The author would like to thank the volunteers who helped validate the dataset labels, Thierry Haddad for providing images of household and recycling center waste, and all DWaste users who submitted images.

\end{ack}

\bibliographystyle{unsrt}
\bibliography{references}

\end{document}